\documentclass[10pt,journal,letterpaper,compsoc]{IEEEtran}
\ifCLASSOPTIONcompsoc
  \usepackage[nocompress]{cite}
\else
  \usepackage{cite}
\fi
\ifCLASSINFOpdf
\else
  \usepackage[dvips]{graphicx}
  \graphicspath{{eps/}}
\fi
\usepackage{algorithmic}
\usepackage{algorithm}
\usepackage{amssymb}
\usepackage{multirow}
\usepackage{siunitx}

\usepackage{array}

\usepackage{amsfonts}
\usepackage{graphicx}
\usepackage{epsfig}
\usepackage{subfigure}
\usepackage{color}

\begin{document}
%
\title{Revealing Cluster Structure of Graph by Path Following Replicator Dynamic}
%
%
%
%

\author{Hairong~Liu, Longin~Jan~Latecki,~\IEEEmembership{Senior Member,~IEEE}, Shuicheng~Yan,~\IEEEmembership{Senior Member,~IEEE}%
\IEEEcompsocitemizethanks{\IEEEcompsocthanksitem
Hairong Liu is with the Dept. of Mechanical Engineering, Purdue University, USA.
\protect\\
E-mail:lhrbss@gmail.com
\protect\\
Longin Jan Latecki is with Dept. of Computer and Information Sciences, Temple University, USA.
\protect\\
E-mail: latecki@temple.edu
\protect\\
Shuicheng Yan is with the Dept. of Electrical and Computer Engineering, National University of Singapore, Singapore.
\protect\\
E-mail: eleyans@nus.edu.sg
}
}

\IEEEcompsoctitleabstractindextext{%
\begin{abstract}
In this paper, we propose a path following replicator dynamic, and investigate its potentials in uncovering the underlying cluster structure of a graph. The proposed dynamic is a generalization of the discrete replicator dynamic, which evolves to a local maximum of the optimization problem $\max f(\mathbf{x})=\mathbf{x}^TW\mathbf{x}, \mathbf{x}\in \Delta$ ($W$ is a non-negative square matrix and $\Delta$ is a simplex), and has been widely used to stimulate the evolution process of animal behavior. The discrete replicator dynamic has been successfully used to extract dense clusters of graphs; however, it is often sensitive to the degree distribution of a graph, and usually biased by vertices with large degrees, thus may fail to detect the densest cluster. To overcome this problem, we introduce a dynamic parameter $\varepsilon$, called path parameter, into the evolution process. That is, we successively solve a series of optimization problems, $\max f(\mathbf{x})=\mathbf{x}^TW\mathbf{x}, \mathbf{x}\in \Delta_{\varepsilon}$, where $\Delta_{\varepsilon}=\{\mathbf{x}|\sum_ix_i=1,x_i\in[0,\varepsilon]\}$. The path parameter $\varepsilon$ can be interpreted as the maximal possible probability of a current cluster containing a vertex, and it monotonically increases as evolution process proceeds. By limiting the maximal probability, the phenomenon of some vertices dominating the early stage of evolution process is suppressed, thus making evolution process more robust. To solve the optimization problem with a fixed $\varepsilon$, we propose an efficient fixed point algorithm. Intuitively, the proposed dynamic follows the solution path of the optimization problems $\max f(\mathbf{x})=\mathbf{x}^TW\mathbf{x}, \mathbf{x}\in \Delta_{\varepsilon}$, with gradually expanding domain $\Delta_{\varepsilon}$. The key properties of the proposed path following replicator dynamic are: $1$) its probability to evolve to the most significant cluster of graph is much higher than discrete replicator dynamic, $2$) the path parameter $\varepsilon$ offers us a tool to control the evolution process, and we can use it to simultaneously obtain dense subgraphs of various specified sizes, and $3$) the evolution process is essentially a shrink process of high-density regions, thus reveals the underlying cluster structure of graph. The time complexity of the path following replicator dynamic is only linear in the number of edges of a graph, thus it can analyze graphs with millions of vertices and tens of millions of edges on a common PC in a few minutes. Besides, it can be naturally generalized to hypergraph and graph with edges of different orders, where $f(\mathbf{x})$ becomes a polynomial function. We apply it to four important problems: maximum clique problem, densest $k$-subgraph problem, structure fitting, and discovery of high-density regions. The extensive experimental results clearly demonstrate its advantages, in terms of robustness, scalability and flexility.
\end{abstract}

}

\maketitle

\IEEEdisplaynotcompsoctitleabstractindextext

%
\IEEEpeerreviewmaketitle

\section{Introduction}
\IEEEPARstart{R}eplicator dynamic \cite{bomze1983lotka} is a deterministic monotone non-linear game dynamic used in evolutionary game theory \cite{weibull1997evolutionary}, a field which models the evolution process of animal behavior using game theory. Considering a large population of individuals which compete for a particular limited resource, each individual can choose one strategy from a set of predefined strategies. Let $I=\{1,\ldots,n\}$ be the set of pure strategies and let $x_i(t)$ be the probability of the population members playing strategy $i$ at time $t$. The state of the system at time $t$ is represented by a vector $\mathbf{x}(t)=(x_1(t),\ldots,x_n(t))^T$. Let $\mathbf{W}=(w_{ij})$ be the $n\times n$ payoff matrix. Specially, for each pair of strategies $i,j\in I$, $w_{ij}$ represents the payoff of an individual playing strategy $i$ against an opponent playing strategy $j$. Without loss of generality, we shall assume $\mathbf{W}$ is nonnegative, i.e., $w_{ij}\geq 0$ for all $i,j\in I$. The discrete replicator dynamic (DRD) can be expressed in the following form \cite{bomze1983lotka}:
\begin{equation}\label{dre}
x_i(t+1) = \frac{x_i(t)(\mathbf{W}\mathbf{x}(t))_i}{\mathbf{x}^T(t)\mathbf{W}\mathbf{x}(t)}.
\end{equation}

It is well known that (\ref{dre}) is a growth transformation for the following optimization problem \cite{baum1967inequality}:
\begin{eqnarray}\label{optproblem1}
\max_\mathbf{x} f(\mathbf{x})=\mathbf{x}^T\mathbf{W}\mathbf{x}, \mathbf{x}\in\Delta,
\end{eqnarray}
where $\Delta=\{\mathbf{x}|\sum_ix_i=1,x_i\in[0,1]\}$ is a simplex. In fact, the sequence $\{\mathbf{x}(t)\}$ converges to a local maximum of (\ref{optproblem1}) along an increasing trajectory \cite{baum1967inequality}.

For each graph $G=(V,E,W)$, we can regard it as a game, with each vertex $v\in V$ representing a strategy, and the weighted adjacency matrix of graph $G$ being the payoff matrix $\mathbf{W}$. Then the evolution process (\ref{dre}) can be interpreted as a cluster extraction process \cite{Masg2009}. $\mathbf{x}$ is a soft indicator vector of the cluster $C_{\mathbf{x}}$, with $x_i$ being the probability of the cluster $C_{\mathbf{x}}$ to contain vertex $v_i$. The objective function $f(\mathbf{x})$ in (\ref{optproblem1}) is a measure of compactness of the cluster $C_{\mathbf{x}}$, and each local maximum of (\ref{optproblem1}) represents a potential dense cluster on $G$ \cite{liu2010common,liu2010robust1}. Many previous works describe the relations between local maxima of (\ref{optproblem1}) and clusters on $G$ \cite{liu2010robust1,motzkin1965maxima,pavan2007dominant}. For example, the well-known Motzkin-Straus theorem \cite{motzkin1965maxima} states that for unweighted graph, the global maximum of (\ref{optproblem1}) corresponds to the maximum clique on $G$.



Despite its success in cluster extraction, discrete replicator dynamic is limited in several aspects. First, although it evolves to a maximum of (\ref{optproblem1}), this maximum may be not the global maximum of (\ref{optproblem1}). In fact, the evolution process (\ref{dre}) is very sensitive to the degree distribution of graph $G$, and the vertices with high degrees usually bias the evolution process \cite{motzkin1965maxima}. Second, the clusters corresponding to the local maxima of (\ref{optproblem1}) are often extremely compact, such as the cliques on unweighted graphs \cite{bomze1999maximum}. In some applications, this may be a problem, since the desired clusters are not so compact. In such situations, we may want to control the sizes of detected clusters. Third, there may be multiple dense clusters on graph, but discrete replicator dynamic only reveals one of them. For example, in a social network graph, there are usually multiple communities. These communities are relatively compact, but not as compact as cliques. Ideally, we need an algorithm to automatically reveal all these communities.

To fulfil these practical needs, in this paper, we propose a new replicator dynamic, called path following replicator dynamic (PFRD). Our proposed dynamic is based on the following observation: the sensitivity of the discrete replicator dynamic (\ref{dre}) to the degree distribution of graph $G$ is mainly caused by the fact that vertices with high degrees often have too large values in $\mathbf{x}$ at the early stage of evolution process. As a consequence, these vertices may dominate the replicator process. For example, in the maximum clique problem, if the initialization is $\mathbf{x}(0)=\{\frac{1}{n},\ldots,\frac{1}{n}\}$, which is commonly used \cite{bomze1999maximum}, then $\mathbf{x}(1)=\{\frac{d_1}{d_{\Sigma}},\ldots,\frac{d_n}{d_{\Sigma}}\}$, where $d_i$ is the degree of $v_i$ and $d_{\Sigma}=\sum_{i=1}^nd_i$. If a vertex, say $v_i$, has a very large degree, then $\mathbf{x}_1(i)$ will be very large, and the replicator process will be greatly biased and most likely to converge to a clique containing $v_i$. If $v_i$ does not belong to the maximum clique, then the discrete replicator dynamic (\ref{dre}) converges to a local maximum, not the global maximum of (\ref{optproblem1}).

We propose to overcome this problem by introducing a parameter $\varepsilon$, called path parameter, into the evolution process. Mathematically speaking, we sucessively solve a series of optimization problems in the following form, with monotonically increasing $\varepsilon$:
\begin{eqnarray}\label{optproblem2}
\max_\mathbf{x} f(\mathbf{x})=\mathbf{x}^TW\mathbf{x}, \mathbf{x}\in \Delta_{\varepsilon}.
\end{eqnarray}
where $\Delta_{\varepsilon}=\{\mathbf{x}|\sum_{i=1}^nx_i=1,x_i\in[0,\varepsilon]\}$ is a subset of the simplex $\Delta$.

Due to the constraint $\sum_{i=1}^nx_i=1$, the minimal value of $\varepsilon$ is $\frac{1}{n}$. Theoretically, $\varepsilon$ can continuously increase from $\frac{1}{n}$ to $1$. In our implementation, we samples $m$ discrete values within the range $[\frac{1}{n},1]$, which form a set $\Phi=\{\varepsilon_1,\ldots,\varepsilon_m\}$, with $\varepsilon_1=\frac{1}{n}$, $\varepsilon_m\leq 1$, and $\varepsilon_i<\varepsilon_{i+1}$ for all $i=1,\ldots,m-1$. When $\varepsilon=\frac{1}{n}$, the maximizer of (\ref{optproblem2}) is $x=\{\frac{1}{n},\ldots,\frac{1}{n}\}$. In our approach, the local maximizer of (\ref{optproblem2}) at $\varepsilon=\varepsilon_i$ serves as the initialization of (\ref{optproblem2}) at $\varepsilon=\varepsilon_{i+1}$. In this way, $x$ follows the solution path of (\ref{optproblem2}) as $\varepsilon$ increases from $\frac{1}{n}$ to $\varepsilon_m$. The choice of $\varepsilon_m$ depends on the applications. When $\varepsilon_m=1$, we get a local maximizer of (\ref{optproblem1}); however, due to the path parameter $\varepsilon$, this local maximizer is more likely to be the global maximizer of (\ref{optproblem1}), as explained later and verified by experiments.

The proposed approach is a direct generalization of discrete replicator dynamic, and it has several advantages. First, since $\varepsilon$ gradually increases from $\frac{1}{n}$ to $\varepsilon_m$, at early stage of evolution process, no vertex has large values. Thus, the evolution process is not sensitive to the degree distribution and mainly depends on the overall structure of graph. When $\varepsilon_m=1$, the evolution process has much higher probability to converge to the global optimum of (\ref{optproblem1}), compared with discrete replicator dynamic. From another prospective, the path parameter $\varepsilon$ prohibits that some components of $\mathbf{x}$ suddenly increase from small values to very large values, which is also reasonable for simulation of animal behavior. As a whole, animals change their behaviors gradually, not suddenly. Thus, the proposed dynamic may better stimulate the evolution process of animal behaviors. Second, the parameter $\varepsilon$ offers us a flexible tool to control the evolution process. Since $x_i\leq \varepsilon$ and $\sum_{i=1}^nx_i = 1$, there are at least $\lceil\frac{1}{\varepsilon}\rceil$ nonzero components in $\mathbf{x}$, where $\lceil\frac{1}{\varepsilon}\rceil$ represents the smallest integer larger than or equal to $\frac{1}{\varepsilon}$. In other words, the cluster $C_{\mathbf{x}}$ contains at least $\lceil\frac{1}{\varepsilon}\rceil$ vertices. Hence, we can control the size of obtained clusters. For example, if we want to get a dense cluster with $k$ vertices, then we can simply set $\varepsilon_m=\frac{1}{k}$. A typical application of this property is the densest $k$-subgraph problem (D$k$S) \cite{feige2001dense}. Note that some components of $\mathbf{x}$ may be very small, thus, in our implementation, we use a threshold $\delta_1$ to judge whether a vertex belongs to $C_{\mathbf{x}}$ or not. Specifically, we define $C_{\mathbf{x}}=\{v_i|x_i>\delta_1\}$, where $\delta_1$ is a small constant whose value is specifically decided for different applications. Third, since the evolution process mainly depends on the overall structure of graph, it can robustly reveal the underlying cluster structure. In fact, the evolution process can be regarded as a gradual simplification process of graph $G$, and as $\varepsilon$ increases, vertices which have relatively weak connections with current graph shall be dropped. Clearly, this is essentially a shrink process of high-density regions in the data, and the dropped vertices can be considered as outliers. Note that discrete replicator dynamic does not have such properties, since it is sensitive to degree distribution. Fourth, the proposed approach is very efficient, with linear time complexity in the number of edges. Thus, we can efficiently analyze the cluster structure of very large graphs, such as network graphs.

\begin{figure*}[t]
\centering
\includegraphics[width=1\linewidth]{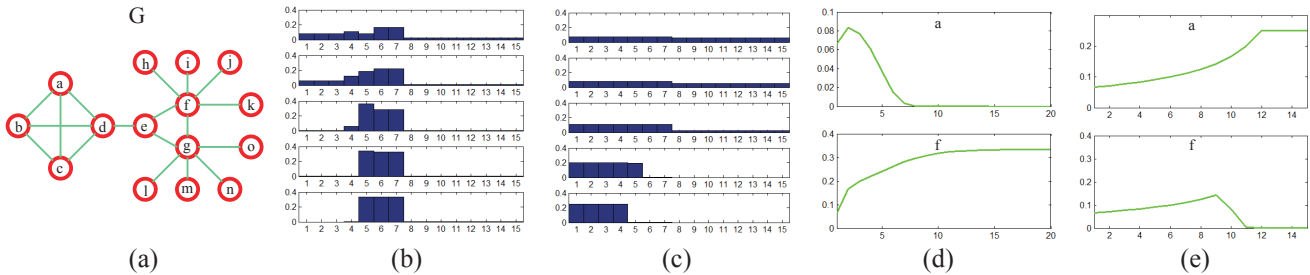}
\caption{Evolution processes of discrete replicator dynamic and the proposed path following replicator dynamic. (a) A graph with $15$ vertices, with a clique of size $4$, \{a,b,c,d\}, and a clique of size $3$, \{e,f,g\}. (b) Evolution process of discrete replicator dynamic. Vertices f and g have relatively large degrees, and thus have relative large values $\mathbf{x}_6$ and $\mathbf{x}_7$ at the beginning of evolution process. The process wrongly evolves to a local maximum of (\ref{optproblem1}), which corresponds to the clique \{e,f,g\}. (c) Evolution process of path following replicator dynamic. The components of $\mathbf{x}$ are constrained by the path parameter, thus no vertex can dominate the evolution process at early stages, and the process correctly converges to the global maximum of (\ref{optproblem1}), which corresponds to the maximum clique \{a,b,c,d\}. The evolution process of our approach also clearly reveals the cluster structure of this graph: the vertices h, i, j, k, l, m, n and o disappear first, then the vertices in the clique \{e,f,g\}, finally only the vertices in the maximal clique remains. For comparison, the evolution processes of components corresponding to nodes a and f in discrete replicator dynamic and our approach are shown in (d) and (e), respectively.
}
\end{figure*}

Fig. $1$ demonstrates the evolution processes of both discrete replicator dynamic and path following replicator dynamic on an unweighted graph $G$.
The components of $\mathbf{x}$ is in accordance with the lexicographical order of the vertices, that is, $x_1$ represents a, $x_{15}$ represents o, and so on. According to the Motzkin-Straus theorem, the global maximum of (\ref{optproblem1}) corresponds to the maximum clique, \{a,b,c,d\}. However, due to sensitivity to degree distribution, the discrete replicator dynamic wrongly converges to a local maximum of (\ref{optproblem1}) corresponding to the clique \{e,f,g\}. In comparison, our algorithm evolves to the global maximum of (\ref{optproblem1}) and correctly finds the maximum clique. In the evolution process of discrete replicator dynamic, $x_6$ and $x_7$ have relatively large values after the first iteration since f and g have more neighbors. Hence, the evolution process is biased by f and g\footnote{Such phenomenon is more obvious and dominant on large graphs.}. The value of $x_6$ as a function of $t$ is shown in (d). Obviously, it increases quickly in the first few iterations. In the path following replicator dynamic, all components of $\mathbf{x}$ change more smoothly, e.g. $x_1$ and $x_6$ shown in (e).

The rest of the paper is organized as follows. We first review the related work in Section $2$. In Section $3$, a fixed point method is introduced to efficiently solve (\ref{optproblem2}), and the sampling strategy for path parameter is also discussed. The experimental results on four problems, namely, maximum clique problem, densest $k$-subgraph problem, structure fitting and discovery of high-density regions, are demonstrated in Section $4$. In Section $5$, the conclusive remarks are made.

\section{Related Work}
Cluster analysis is a basic problem in various disciplines \cite{xu2005survey}, such as pattern recognition, data mining and computer vision, and a huge number of such methods have been proposed. It is beyond the scope of this paper to list all of them, therefore, we focus on methods closely related to ours.

The discrete replicator dynamic has been used for cluster extraction for a very long time, probably dating back to the well-known work of Motzkin and Straus \cite{motzkin1965maxima}, which relates the global maximum of (\ref{optproblem1}) with the maximum clique of graph. The maximum clique problem \cite{bomze1999maximum} is a very fundamental problem in computer science, but it is NP-complete \cite{motzkin1965maxima}. According to the Motzkin-Straus theorem, we can get the maximum clique by solving (\ref{optproblem1}). In \cite{pavan2007dominant}, the discrete replicator dynamic has been used to extract cluster on weighted graph. The obtained cluster, called dominant set, is a generalization of the concept of maximal clique. To efficiently enumerate dense clusters on graph, a fast algorithm has been proposed in \cite{liu2010robust1}, which evolves to a dense subgraph by iteratively shrinking and expansion. In the shrink stage of this method, the discrete replicator dynamic is used to extract dense clusters. The clustering methods based on replicator dynamic have been successfully applied to many tasks, such as image segmentation \cite{pavan2007dominant}, point set and image matching \cite{liu2010robust1,liu2010common,pelillo2002matching,MHs1999,Masg2009} and stereo correspondence \cite{horaud1989stereo}. All these methods rely on discrete replicator dynamic to get the global optimum of (\ref{optproblem1}); however, discrete replicator dynamic (\ref{dre}) is very sensitive to the degree distribution of graph $G$, and usually does not evolve to the global maximum of (\ref{optproblem1}). Since our approach has a much higher probability to evolve to the global maximum of (\ref{optproblem1}) than discrete replicator dynamic, it is more suitable for these applications.

Methods which try to extract dense clusters of certain sizes are also related to the proposed approach. For example, the methods to solve the densest $k$-subgraph problem (D$k$S) \cite{feige2001dense,roupin2008deterministic}. In hypergraph clustering, Liu et al. \cite{liu2010affinity} proposed a method to control the least size of extracted clusters, and showed state-of-the-art results. Utilizing the fact that dense clusters are very robust, they also generalized $k$NN and proposed an alternative, $k$DN \cite{denseneigh2012}. $k$DN is in fact a dense cluster of size $k$ which has a strong connection with an object. Obviously, our approach can be used to solve these problems. In fact, our approach is more robust. In our approach, the parameter $\varepsilon$ controls the least size of obtained clusters. Moreover, our approach can simultaneously and robustly find dense clusters of different sizes in one run of evolution process.

Since dense subgraphs correspond to high-density regions in the data, the evolution process of path following replicator dynamic can be considered as a
shrink process of high-density regions. The task of estimating high-density regions from data samples is a fundamental problem in a number of works, such as outlier detection and cluster analysis \cite{scholkopf2001estimating,munoz2006estimation}. The advantage of our method for this task is that our method can gradually reveal the landscape of multiple high-density regions of various shape at different scales. High density regions usually represent modes of data, and in this sense, our method is also closely related to mode-finding methods, such as mean shift \cite{comaniciu2002mean}.

\section{Algorithm}
The central part of path following replicator dynamic is to efficiently solve (\ref{optproblem2}). We first analyze the properties of the solution, then present our algorithm.

\subsection{Properties of Local Maximizers}
In \cite{liu2010affinity}, the properties of local maximizers of (\ref{optproblem2}) have been analyzed. Here we give a brief summary.

By adding Lagrangian multipliers $\lambda$, $\mu_1,\cdots,\mu_n$, $\mu_i\geq 0$ and $\nu_1,\cdots,\nu_n$, $\nu_i\geq 0$ for all $i=1,\cdots,n$, we can obtain the Lagrangian function of (\ref{optproblem2}):
\begin{equation}
L(\mathbf{x},\lambda,\mathbf{\mu}) = f(x)-\lambda(\sum_{i=1}^nx_i-1)+\sum_{i=1}^n\mu_ix_i+\sum_{i=1}^n\nu_i(\varepsilon-x_i).
\end{equation}
According to Karush-Kuhn-Tucker (KKT) condition \cite{kuhn1951nonlinear}, if $\mathbf{x}^*$ is a local maximizer of (\ref{optproblem2}), then
\begin{equation}
\left \{
\begin{array}{l}
\frac{\partial f}{\partial x_i}(\mathbf{x})-\lambda+\mu_i-\nu_i=0, \mbox{ }i=1,\cdots,n,\\
\sum_{i=1}^n\mu_ix^*_i=0,\\
\sum_{i=1}^n\nu_i(\varepsilon-x^*_i)=0.
\end{array}
\right.
\end{equation}

\noindent$\frac{\partial f}{\partial x_i}(\mathbf{x})$ represents the partial derivative of $f(\mathbf{x})$ with respect to $x_i$, and the partial derivatives of $f(\mathbf{x})$ with respect to all components of $\mathbf{x}$ form a vector $g(\mathbf{x})=\frac{\partial f}{\partial \mathbf{x}}(\mathbf{x})$. Since $x^*_i$, $\mu_i$ and $\nu_i$ are nonnegative for all $i=1,\cdots,n$, $\sum_{i=1}^n\mu_ix^*_i=0$ is equivalent to saying that if $x^*_i>0$, then $\mu_i=0$, and $\sum_{i=1}^n\nu_i(\varepsilon-x^*_i)=0$ is equivalent to saying that if $x^*_i<\varepsilon$, then $\nu_i=0$. Hence, based on simple algebraic calculations, the KKT conditions can be rewritten in the following form:
\begin{equation}\label{kktf}
\frac{\partial f}{\partial x_i}(\mathbf{x})\left\{
\begin{array}{cc}
         \leq \lambda, & x^*_i = 0; \\
         = \lambda, & 0<x^*_i<\varepsilon; \\
         \geq \lambda, & x^*_i=\varepsilon.
 \end{array}
                         \right.
\end{equation}

Any point satisfying the KKT condition (\ref{kktf}) is called a \emph{KKT point}. Since KKT condition is a necessary condition, a local maximizer of (\ref{kktf}) is also a KKT point. However, a KKT point may be not a local maximizer of (\ref{kktf}).

The KKT condition (\ref{kktf}) is important for our algorithmic design. It has an intuitive geometric meaning: the partial derivatives with respect to all variables in the range $(0,\varepsilon)$ have the same value $\lambda$, the partial derivatives with respect to variables having value $0$ should be not larger than $\lambda$, and the partial derivatives with respect to all variables having value $\varepsilon$ should be not smaller than $\lambda$.

\subsection{Truncated Simplex Projection}
In this section, we propose an efficient algorithm to solve (\ref{optproblem2}). The discrete replicator iteration (\ref{dre}) can be rewritten into the following form:
\begin{equation}\label{dri}
\mathbf{x}(t+1) = \mbox{proj}_{\Delta}(\mathbf{x}(t)\odot g(\mathbf{x}(t))),
\end{equation}
where $\odot$ stands for element-wise multiplication, and $\mbox{proj}_{\Delta}(\mathbf{y})$ is a projection, which projects a nonnegative vector $\mathbf{y}$ onto the simplex $\Delta=\{\mathbf{x}|x_i\geq 0, \sum_{i=1}^nx_i=1\}$ by $\ell_1$ normalization. That is,
\begin{equation}
\mbox{proj}_{\Delta}(\mathbf{y})_i=\frac{y_i}{\sum_{j=1}^ny_j}.
\end{equation}
This projection is called \textbf{simplex projection}, and its main characteristic is: all components of $\mathbf{y}$ scale uniformly.

For the problem (\ref{optproblem2}), the feasible region of $\mathbf{x}$ is $\Delta_{\varepsilon}=\{\mathbf{x}|x_i\in [0,\varepsilon], \sum_{i=1}^nx_i=1\}\}$, a subset of $\Delta$. Obviously, simplex projection cannot be used here, since some components may exceed $\varepsilon$ after projection. To overcome this problem, a natural idea is: if some components exceed $\varepsilon$ after projection, then we set their values to be $\varepsilon$ and scale other components uniformly, this process iterates until no component is larger than $\varepsilon$ after projection. The new projection $\mbox{proj}_{\Delta_{\varepsilon}}(\mathbf{y})$, called \textbf{truncated simplex projection}, is a direct generalization of the simplex projection.

Suppose $U$ contains all components of $\mathbf{y}$ whose values should be set to $\varepsilon$ after projection and $V=I/U$, then the truncated simplex projection can be mathematically described in the following form:
\begin{equation}\label{po}
\mbox{proj}_{\Delta_{\varepsilon}}(\mathbf{y})_i=\left\{
\begin{array}{cc}
         \varepsilon, & i\in U,\\
         \frac{(1-|U|\varepsilon)}{\sum_{j\in V} y_j}y_i, & i\in V,
 \end{array}
\right.
\end{equation}
where $|U|$ is the cardinality of set $U$. Obviously, $U$ should satisfy two criteria: 1) the components of $\mathbf{y}$ in $U$ are larger than the components of $\mathbf{y}$ in $V$, and 2) no element in $U$ can be moved to $V$. In (\ref{po}), all components in $U$ are set to $\varepsilon$, and other components are scaled to make the sum of all components equal to $1$, thus the scaling factor is $\frac{(1-|U|\varepsilon)}{\sum_{j\in V} y_j}$.

The algorithm to compute the truncated simplex projection of a nonnegative vector $\mathbf{y}$ is summarized in Alg. $1$. The critical part is to determine the set $U$. Since the components of $\mathbf{y}$ in $U$ are larger than the components in $V$, we first sort all components of $\mathbf{y}$ in descending order, that is, $\{y_{s_1},\ldots,y_{s_n}\}$, with $y_{s_i}\geq y_{s_{i+1}}$ for all $i=1,\ldots,n-1$, and then check them one by one, from $y_{s_1}$ to $y_{s_n}$. When we check the $i$-th component $y_{s_i}$, the first $i-1$ components are all in $U$ and the values of these components should be set to $\varepsilon$ after projection, then the sum of other $n-i+1$ components, from the $i$-th component to the $n$-th component, should be $1-(i-1)\varepsilon$ after projection. Since the sum of these $n-i+1$ components before projection is $z=\sum_{j=i}^ny_{s_j}$, then the scale factor is $\frac{1-(i-1)\varepsilon}{z}$. If the value of the $i$-th component is smaller than $\varepsilon$ after projection, then we have already found all elements in $U$; otherwise, we put the $i$-th component in $U$ and check next component $y_{s_{i+1}}$.

\begin{algorithm}[tb]
   \caption{Truncated Simplex Projection  $\mbox{proj}_{\Delta_{\varepsilon}}(\mathbf{y})$}
   \label{alg:gs}
\begin{algorithmic}[1]
   \STATE {\bfseries Input:} The vector $\mathbf{y}$ and the path parameter $\varepsilon$.
   \STATE Sort the components of $\mathbf{y}$ in descending order, $\{y_{s_1},\ldots,y_{s_n}\}$.
   \STATE Compute $z=\sum_{i=1}^ny_i$.
   \STATE Set $U=\emptyset$.
   \FOR{$i=1,\ldots,n$}
   \STATE Compute $\chi=\frac{(1-(i-1)\varepsilon)y_{s_i}}{z}$. If $\chi\geq \varepsilon$, set $U=U\cup\{s_i\}$ and $z=z-y_{s_i}$; Otherwise, break.
   \ENDFOR
   \STATE Set $V=I/U$. For all $i\in U$, set $\mbox{proj}_{\Delta_{\varepsilon}}(\mathbf{y})_i=\varepsilon$. For all $i\in V$, set $\mbox{proj}_{\Delta_{\varepsilon}}(\mathbf{y})_i=\frac{y_i(1-|U|\varepsilon)}{z}$.
   \STATE{\bfseries Output:} $\mathbf{y'}=\mbox{proj}_{\Delta_{\varepsilon}}(\mathbf{y})$.
\end{algorithmic}
\end{algorithm}

It is easy to verify that: $1$) $\mbox{proj}_{\Delta_{\varepsilon}}(\mathbf{y})$ is a nonnegative vector, and the sum of all its components is equal to $1$, $2$) $\mbox{proj}_{\Delta_{\varepsilon}}(\mathbf{y})_i\leq \varepsilon$ for all $i=1,\ldots,n$, and $3$) if $y_i\geq y_j$, then $\mbox{proj}_{\Delta_{\varepsilon}}(\mathbf{y})_i\geq \mbox{proj}_{\Delta_{\varepsilon}}(\mathbf{y})_j$. Clearly, the simplex projection $\mbox{proj}_{\Delta}(\mathbf{y})$ used in discrete replicator dynamic is in fact a special case of the truncated simplex projection $\mbox{proj}_{\Delta_{\varepsilon}}(\mathbf{y})$ at $\varepsilon=1$.

\subsection{Fixed Point Iteration}
Based on truncated simplex projection, we can define an iteration similar to (\ref{dri}):
\begin{equation}\label{fpit}
\mathbf{x}(t+1) = \mbox{proj}_{\Delta_{\varepsilon}}(\mathbf{x}(t)\odot g(\mathbf{x}(t))).
\end{equation}
Obviously, this iteration ensures that $\mathbf{x}$ always stays in the region $\Delta_{\varepsilon}$.

For simplicity of notation, we define an operator $P_{\varepsilon}(\mathbf{x})=\mbox{proj}_{\Delta_{\varepsilon}}(\mathbf{x}\odot g(\mathbf{x}))$. Then the iteration (\ref{fpit}) can be simply expressed in the following way:
\begin{equation}\label{fpi}
\mathbf{x}(t+1) = P_{\varepsilon}(\mathbf{x}(t)).
\end{equation}
From an initialization $\mathbf{x}(0)$, we can repeat this iteration until a fixed point of $P_{\varepsilon}(\mathbf{x})$ is obtained. The following two theorems build the relation between the fixed points of the operator $P_{\varepsilon}(\mathbf{x})$ and the KKT points of (\ref{optproblem2}), and they form the theoretic foundation of path following replicator dynamic.

\vspace{5pt}
\noindent \textbf{Theorem 1}. The KKT point $\mathbf{\tilde{x}}$ of (\ref{optproblem2}) is a fixed point of the operator $P_{\varepsilon}(\mathbf{x})$.
\vspace{5pt}

\noindent \textbf{Proof}: Since $\mathbf{\tilde{x}}$ is a KKT point of (\ref{optproblem2}), according to the KKT condition (\ref{kktf}), we get:
\begin{equation}
g_i(\mathbf{\tilde{x}})\left\{
\begin{array}{cc}
         \leq \lambda, & \tilde{x}_i =0;\\
         = \lambda, & 0<\tilde{x}_i<\varepsilon; \\
         \geq \lambda, & \tilde{x}_i=\varepsilon.
 \end{array}
                         \right.
\end{equation}
Let $U_1=\{i|\tilde{x}_i=\varepsilon\}$, $U_2=\{i|\tilde{x}_i\in (0,\varepsilon)\}$, $U_3=\{i|\tilde{x}_i=0\}$ and $\mathbf{x'} = P_{\varepsilon}(\mathbf{\tilde{x}})$. We only need to prove that $U=U_1$ in Alg. $1$. This is because when $U=U_1$, we can get: 1) when $i\in U_1$, according to Alg. $1$,  $x'_i=\varepsilon$, 2) when $i\in U_3$, obviously, $x'_i=0$, and 3) when $i\in U_2$, since $g_i(\mathbf{\tilde{x}})=\lambda$ and the components in $U_2$ scale uniformly, then $x'_i=\tilde{x}_i$. Thus, when $U=U_1$, $\mathbf{x'}=\mathbf{\tilde{x}}$ and $\mathbf{\tilde{x}}$ is a fixed point of $P_{\varepsilon}(\mathbf{x})$.

Now we prove that $U=U_1$. After sorting the components of $\mathbf{y}=\mathbf{\tilde{x}}\odot g(\mathbf{\tilde{x}})$ in descending order, all elements in $U_1$ are at the front, followed by the elements in $U_2$, and by the elements in $U_3$. According to Alg. $1$, we need to prove two inequalities:
\begin{eqnarray}
\frac{y_{s_{|U_1|}}(1-(|U_1|-1)\varepsilon)}{\sum_{i=|U_1|}^ny_{s_i}}\geq \varepsilon, \label{cond1}\\
\frac{y_{s_{|U_1|+1}}(1-|U_1|\varepsilon)}{\sum_{i=|U_1|+1}^ny_{s_i}}<\varepsilon. \label{cond2}
\end{eqnarray}
The first inequality ensures that the $|U_1|$-th largest component of $\mathbf{y}$ is in $U$, and the second inequality ensures that the $|U_1|+1$-th largest component of $\mathbf{y}$ is not in $U$.

Since $1-|U_1|\varepsilon=\sum_{i=|U_1|+1}^n\tilde{x}_{s_i}$ and $\forall i>|U_1|, y_{s_i}=\lambda \tilde{x}_{s_i}$, we get:
\begin{equation}
\frac{y_{s_{|U_1|+1}}(1-|U_1|\varepsilon)}{\sum_{i=|U_1|+1}^ny_{s_i}}=\frac{\lambda\tilde{x}_{s_{|U_1|+1}}(1-|U_1|\varepsilon)}{\lambda\sum_{i=|U_1|+1}^n\tilde{x}_{s_i}}=\tilde{x}_{s_{|U_1|+1}}<\varepsilon. \end{equation}
Thus, (\ref{cond2}) holds.

For (\ref{cond1}), from $y_{s_i}=x_{s_i} g_{s_i}(\mathbf{\tilde{x}})$ and $x_{s_|U_1|}=\varepsilon$, we get:
\begin{eqnarray}
&&\frac{y_{s_{|U_1|}}(1-(|U_1|-1)\varepsilon)}{\sum_{i=|U_1|}^ny_{s_i}}\geq \varepsilon \nonumber\\
&\Longleftrightarrow & g_{s_{|U_1|}}(\mathbf{\tilde{x}})\geq \frac{\sum_{i=|U_1|+1}^n x_{s_i}g_{s_i}(\mathbf{\tilde{x}})}{1-|U_1|\varepsilon}=\lambda, \nonumber
\end{eqnarray}
which is true according to the KKT condition (\ref{kktf}). $\blacksquare$

When $\mathbf{\tilde{x}}$ is a fixed point of the operator $P_{\varepsilon}(\mathbf{x})$, is it a KKT point of (\ref{optproblem2})? This question is complicated, and the following theorem partially answers this question.

\vspace{5pt}
\noindent \textbf{Theorem 2}. If $\mathbf{\tilde{x}}$ is a fixed point of the operator $P_{\varepsilon}(\mathbf{x})$, then
we get:
\begin{equation}\label{pkktf}
g_i(\mathbf{\tilde{x}})\left\{
\begin{array}{cc}
         = \lambda, & 0<\tilde{x}_i<\varepsilon; \\
         \geq \lambda, & \tilde{x}_i=\varepsilon.
 \end{array}
                         \right.
\end{equation}
where $\lambda$ is a constant.

\vspace{5pt}
\noindent \textbf{Proof}:
Let $U_1=\{i|\tilde{x}_i=\varepsilon\}$, $U_2=\{i|\tilde{x}_i\in (0,\varepsilon)\}$, $U_3=\{i|\tilde{x}_i=0\}$ and $\mathbf{x'} = P_{\varepsilon}(\mathbf{\tilde{x}})$. Since $\mathbf{\tilde{x}}$ is a fixed point of the operator $P_{\varepsilon}(\mathbf{x})$, $\mathbf{x'}=\mathbf{\tilde{x}}$. According to Alg. $1$, $U=U_1$ and $V=U_2\cup U_3$.

We first prove that $g_i(\mathbf{\tilde{x}})= \lambda$ when $0<\tilde{x}_i<\varepsilon$, that is, when $i\in U_2$.
For any $i,j\in U_2$, since the projection $\mbox{proj}_{\Delta_{\varepsilon}}(\mathbf{y})$ does not change the relative scales of components in $V$, then $\frac{x'_i}{x'_j}=\frac{y_i}{y_j}=\frac{\tilde{x}_ig_i(\mathbf{\tilde{x}})}{\tilde{x}_jg_j(\mathbf{\tilde{x}})}$. Since $\frac{x'_i}{x'_j}=\frac{\tilde{x}_i}{\tilde{x}_j}$, then $g_i(\mathbf{\tilde{x}})=g_j(\mathbf{\tilde{x}})$. That is, the partial derivatives with respect to all variables in $U_2$ are the same, and we denote them by $\lambda$.

Now we prove that $g_i(\mathbf{\tilde{x}})\geq\lambda$ when $\tilde{x}_i=\varepsilon$, that is, when $i\in U_1$.
Suppose $x_i$ is the element in $U_1$ with the smallest partial derivative, according to Alg. $1$, we have:
\begin{equation}
\frac{y_i(1-(|U_1|-1)\varepsilon)}{y_i+\sum_{j\in U_2}y_j}\geq \varepsilon.
\end{equation}
Recall that $y_i=\varepsilon g_i(\mathbf{\tilde{x}})$ and $\sum_{j\in U_2}y_j=(1-|U_1|\varepsilon)\lambda$, then we get:
\begin{eqnarray}
&&\frac{y_i(1-(|U_1|-1)\varepsilon)}{y_i+\sum_{j\in U_2}y_j}\geq \varepsilon \nonumber \\
&\Longrightarrow &y_i(1-(|U_1|-1)\varepsilon)\geq \varepsilon(y_i+\sum_{j\in U_2}y_j) \nonumber \\
&\Longrightarrow &y_i(1-|U_1|\varepsilon)\geq \varepsilon\sum_{j\in U_2}y_j \nonumber\\
&\Longrightarrow &g_i(\mathbf{\tilde{x}})\geq \lambda \nonumber
\end{eqnarray}
Since $x_i$ is the element in $U_1$ with the smallest partial derivative, for any $i\in U_1$, $g_i(\mathbf{\tilde{x}})\geq \lambda$. $\blacksquare$

Note that (\ref{pkktf}) is part of the KKT condition (\ref{kktf}), and the only difference between (\ref{pkktf}) and (\ref{kktf}) is the condition on zero components of $\mathbf{\tilde{x}}$. The KKT condition (\ref{kktf}) requires that the partial derivatives with respect to zero components should be not larger than $\lambda$; however, when $\mathbf{\tilde{x}}$ is a fixed point of the operator $P_{\varepsilon}(\mathbf{x})$, the partial derivatives with respect to zero components can have arbitrary values. This is not surprising, since in the fixed point iteration (\ref{fpi}), when $x_i(t)=0$, then $x_i(t')=0$ for all $t'>t$, no matter how large the partial derivative $g_i(t')$ is. In fact, this is a common characteristic of both (\ref{fpi}) and discrete replicator iteration (\ref{dri}).

In our approach, the initial initialization, that is, the initialization for the optimization problem (\ref{optproblem2}) with $\varepsilon=\varepsilon_1$, is always $\mathbf{x}(0)=\{\frac{1}{n},\ldots,\frac{1}{n}\}$. Theoretically, when graph $G$ does not contain isolated vertices, all components of $\mathbf{x}$ are always positive, although many components approach zeros. Due to arithmetic underflow, some components will become zeros in the iteration process; however, this is because their partial derivatives are very small. Thus, the fixed points of the operator $P_{\varepsilon}(\mathbf{x})$ are usually KKT points of (\ref{optproblem2}). In this evolution process, whether there is an exception or not, that is, whether there is a fixed point of the operator $P_{\varepsilon}(\mathbf{x})$ that is not a KKT point of (\ref{optproblem2}), is an open problem. At least for the discrete replicator dynamic, if the initialization is $\mathbf{x}(0)=\{\frac{1}{n},\ldots,\frac{1}{n}\}$, then it always evolves to a local maximizer of (\ref{optproblem1}), which is a KKT point of (\ref{optproblem1}).

In conclusion, a KKT point of (\ref{optproblem2}) is always a fixed point of the operator $P_{\varepsilon}(\mathbf{x})$, while the fixed point of (\ref{optproblem2}) in our proposed approach is usually a KKT point of (\ref{optproblem2}), although in theory, exception may exist. Besides, according to our experiments, the obtained KKT point is usually a local maximizer of (\ref{optproblem2}). Thus, we can get the solution of (\ref{optproblem2}) with very high probability by the fixed point iteration (\ref{fpi}).

\subsection{Path Following Replicator Dynamic}
Based on the fixed point iteration (\ref{fpi}), we can get the solution of (\ref{optproblem2}) for each $\varepsilon$. The whole algorithm of path following replicator dynamic is summarized in Alg. $2$.

\begin{algorithm}[tb]
   \caption{Path Following Replicator Dynamic}
   \label{alg:gs}
\begin{algorithmic}[1]
   \STATE {\bfseries Input:} $\mathbf{W}$ and $\{\varepsilon_1,\ldots,\varepsilon_m\}$.
   \STATE Set $\mathbf{x}_{\mbox{\tiny init}}=\{\frac{1}{n},\ldots,\frac{1}{n}\}$.
   \FOR{$i=1,\ldots,m$}
   \STATE Set $\varepsilon=\varepsilon_i$, $\mathbf{x}(0)=\mathbf{x}_{\mbox{\tiny init}}$ and $t=0$.
   \REPEAT
   \STATE $t=t+1$;
   \STATE $\mathbf{x}(t)=P_{\varepsilon}(\mathbf{x}(t-1))$;
   \UNTIL{$|\mathbf{x}(t)-\mathbf{x}(t-1)|<\delta_2$}
   \STATE Set $\mathbf{x}_{\mbox{\tiny init}}=\mathbf{x}(t)$ and $\mathbf{\tilde{x}}(i)=\mathbf{x}(t)$.
   \ENDFOR
   \STATE{\bfseries Output:} The solution sequence $\{\mathbf{\tilde{x}}(1),\ldots,\mathbf{\tilde{x}}(m)\}$.
\end{algorithmic}
\end{algorithm}

In Alg. $2$, we use the solution of (\ref{optproblem2}) with $\varepsilon=\varepsilon_i$ as the initialization of the fixed point iteration (\ref{fpi}) with $\varepsilon=\varepsilon_{i+1}$. The fixed point iteration terminates when the change of $\mathbf{x}$ is smaller than a threshold $\delta_2$, which is set to \num{1e-4} in all our experiments. Obviously, the path following replicator dynamic is a direct generalization of discrete replicator dynamic. When the path parameter only has one value, namely, $\varepsilon=1$, the path following replicator dynamic reduces to discrete replicator dynamic.

The output of Alg. $2$ is the solutions of (\ref{optproblem2}) with all values of $\varepsilon$, $\{\mathbf{\tilde{x}}(1),\ldots,\mathbf{\tilde{x}}(m)\}$, which correspond to a series of clusters, $\{C_{\mathbf{\tilde{x}}(1)},C_{\mathbf{\tilde{x}}(2)},\ldots,C_{\mathbf{\tilde{x}}(m)}\}$. Intuitively, in the evolution process, $\mathbf{x}$ follows the solution path of the optimization problem (\ref{optproblem2}) with increasing $\varepsilon$, this is why we called the proposed approach \emph{path following replicator dynamic}.

Each $\mathbf{\tilde{x}}(i)$ represents a dense subgraph of $G$ containing at least $\lceil\frac{1}{\varepsilon_i}\rceil$ vertices. When $\lceil\frac{1}{\varepsilon_i}\rceil$ is large, $C_{\mathbf{\tilde{x}}(i)}$
usually contains about $\lceil\frac{1}{\varepsilon_i}\rceil$ vertices; when $\lceil\frac{1}{\varepsilon_i}\rceil$ is small, $C_{\mathbf{\tilde{x}}(i)}$ may represent a clique (on unweighed graph) or a dominant set (on weighted graph) whose size is larger than $\lceil\frac{1}{\varepsilon_i}\rceil$. In either case, $C_{\mathbf{\tilde{x}}(i)}$ is much denser than other subgraphs with similar sizes, thus represents important pattern underlying the data.
Since $C_{\mathbf{\tilde{x}}(i)}$ is a dense cluster, the sequence $\{C_{\mathbf{\tilde{x}}(1)},C_{\mathbf{\tilde{x}}(2)},\ldots,C_{\mathbf{\tilde{x}}(m)}\}$ can be regarded as gradual simplification of graph $G$, with dense parts of graph $G$ retained. If graph $G$ is constructed from feature points, then the sequence $\{C_{\mathbf{\tilde{x}}(1)},C_{\mathbf{\tilde{x}}(2)},\ldots,C_{\mathbf{\tilde{x}}(m)}\}$ is in fact a shrink process of high-density regions, since dense subgraphs generally correspond to high-density regions \cite{liu2010robust1}. The task of estimating high-density regions from data samples is fundamental in a number of problems, such as outlier detection, cluster analysis and one-class problem\cite{munoz2006estimation,scholkopf2001estimating}. The advantage of our method is that it can gradually reveal multiple high-density regions of various shapes at different scales.

\begin{figure*}[t]
\centering
\includegraphics[width=1\linewidth]{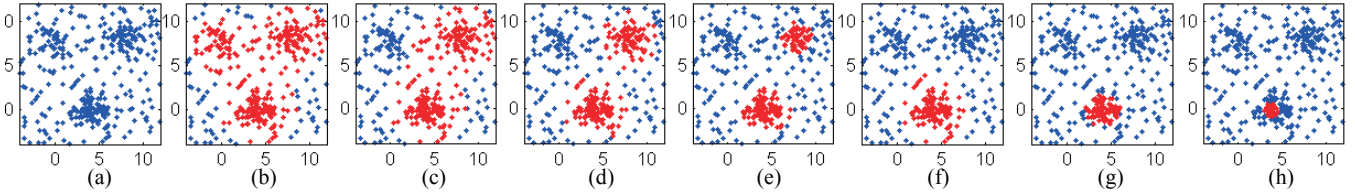}
\caption{Graphical illustration of evolution process of path following replicator dynamic on a planar point cloud. (a) illustrates the point cloud. There are three Gaussian clusters, with $90$, $60$ and $30$ points, respectively. There are also $180$ uniformly distributed outliers. (b)-(h) demonstrate the solution of (\ref{optproblem2}) with $\varepsilon$ being $\frac{1}{300}$, $\frac{1}{240}$, $\frac{1}{200}$, $\frac{1}{160}$, $\frac{1}{120}$, $\frac{1}{80}$ and $1$, respectively.
For each $\mathbf{x}$, the components larger than $\delta_1=\frac{1}{360}$ are illustrated in red, and others are illustrated in blue. Obviously, from this process, all three clusters can be correctly extracted and outliers are elinimated.
}
\end{figure*}

Fig. $2$ demonstrates the evolution process of path following replicator dynamic on a planar point cloud. There are three Gaussian clusters of different densities. The bottom middle cluster is the densest, the top right cluster is the second densest, and the top left cluster is the least dense. Clearly, the evolution process reveals the cluster structure in data. As $\varepsilon$ increases, the detected clusters become more and more compact. Background outliers are dropped first, then points in the least dense cluster are dropped, and then points in the second densest cluster are dropped, finally only a compact subset of points in the densest cluster remain. From this process, we can detect outliers and extract all dense clusters. We emphasize here that all useful information for these tasks is revealed by a single evolution process.

\subsection{Complexity Analysis and Further Speedup}
In the path following replicator dynamic, the basic operation is the iteration $\mathbf{x}(t+1) = P_{\varepsilon}(\mathbf{x}(t))$, which includes three procedures: $1)$ calculation of the partial derivative $g(\mathbf{x})$, $2)$ element-wise multiplication $\mathbf{x}\odot g(\mathbf{x})$, and $3)$ truncated simplex projection. The time complexity of calculating partial derivative is $O(|V|+|E|)$, the time complexity of element-wise multiplication is $O(|V|)$, and the time complexity of truncated simplex projection is $O(|V|\log(|V|))$, since a sort operation is needed. Thus, the time complexity of the iteration $\mathbf{x}(t+1) = P_{\varepsilon}(\mathbf{x}(t))$ is $O(|V|\log(|V|)+|E|)$, which is linear in $|E|$. Obviously, the time complexity of path following replicator dynamic is low. In fact, it can work on graphs with millions of vertices and tens of millions of edges efficiently.

The iteration $\mathbf{x}(t+1) = P_{\varepsilon}(\mathbf{x}(t))$ has a useful property: if $x_i(t)=0$, then $x_i(t')=0$ for all $t'>t$. Thus, if a component of $\mathbf{x}(t)$ becomes zero, we can drop the corresponding vertex and operate on the remaining graph. In path following replicator dynamic, when $G$ has no isolated vertices, theoretically, the components of $\mathbf{x}(t)$ are always positive; however, many components of $\mathbf{x}(t)$ approach $0$ quickly, and then become $0$ due to arithmetic underflow. Thus, we can further accelerate Alg. $2$ by setting very small components of $\mathbf{x}(t)$ to zeros. Specifically, when a component of $\mathbf{x}(t)$ is smaller than a small threshold $\delta_3$, it will be set as zero and the corresponding vertex is dropped. In our experiment, we always set $\delta_3=\num{1e-12}$. Note that even if we do not explicitly use such approximation method, the computer uses it implicitly by means of arithmetic underflow. In our experiments, we only use this approximation method when graph $G$ is very large, such as web graphs, since the approximation has a possibility, although extremely small, to introduce errors.

\subsection{Sampling Strategies for Path Parameter}
Obviously, the path parameter plays a crucial role in path following replicator dynamic, and it is a tool for us to control the evolution process. To obtain the best result, we need to have a good sampling of path parameter. Generally speaking, a larger number of samples of the path parameter, and a better distribution of these samples lead to better results. However, too many samples will render Alg. $2$ inefficient. Compared to the number of samples, the distribution of samples is more important. A sampling with less samples but good distribution is usually better than a sampling with more samples but with bad distribution. Basically, we need to balance between effectiveness and efficiency.

In our experiments, many aspects have been considered to find a good sample strategy for path parameter. First, applications may explicitly require certain number of samples of the path parameter. For example, if we want to find three dense clusters of graph $G$, with sizes are $100$, $200$, $300$, respectively. Then three samples of path parameter, namely, $\frac{1}{300}$, $\frac{1}{200}$ and $\frac{1}{100}$, are needed. Second, the applications may require us to better explore dense subgraphs whose sizes are around a specified size. For example, to detect outliers in a dataset knowing that there are about $10\%$ outliers, we need more samples of path parameter around $\frac{1}{0.9*n}$, where $n$ is the number of data points. Third, to better suppress the possible domination phenomenon, there should be no large intervals in samples. To work out a good sample strategy, we need to comprehensively consider all these aspects.

\subsection{Generalization to Hypergraphs}
The path following replicator dynamic can be easily generalized to hypergraph and graphs with edges of different orders.
From graph to hypergraph, the only difference is $f(\mathbf{x})$, which becomes a polynomial function. Since we do not assume any specific form for $f(\mathbf{x})$ in our algorithmic design, the proposed algorithm can be directly applied to hypergraphs.
Without loss of generality, we also assume the weights of hyperedges are all positive, thus all coefficients of $f(\mathbf{x})$ are positive.


\section{Experiments}
The proposed path following replicator dynamic is a versatile tool for many tasks. First, it is a generalization of discrete replicator dynamic that has a much higher probability to evolve to the global maximum of (\ref{optproblem1}). Thus it can replace the discrete replicator dynamic in many applications, and usually lead to better performance. Second, it can partially control the sizes of extracted clusters, thus have much more flexibility than discrete replicator dynamic. Many applications, such as densest $k$-subgraph problem, constrained clustering \cite{liu2010affinity} and dense neighborhood \cite{denseneigh2012}, require to extract clusters of certain sizes. Third, it essentially reveals the cluster structure of a graph. In the evolution process, outliers are dropped first, then inliers in less compact clusters, finally only vertices in a very compact cluster remain. Thus, it can naturally be used to extract clusters and detect outliers. Note that the main strength of path following replicator dynamic is not to precisely partition the data into clusters, as most clustering methods do \cite{xu2005survey}, but to globally reveal outliers and clusters.

In this section, we apply path following replicator dynamic to four problems, namely, maximum clique problem \cite{motzkin1965maxima}, densest $k$-subgraph problem \cite{feige2001dense}, structure fitting \cite{fischler1981random}, and discovery of high-density regions \cite{munoz2006estimation}. Note that discrete replicator dynamic can be only applied on maximum clique problem, among these four problems.

\subsection{Robustness Testing On Maximum Clique Problem}
In this section, we test the robustness of path following replicator dynamic, that is, its ability to evolve to the global maximum of (\ref{optproblem1}). A good test bed is the maximum clique problem. According to Motzkin-Straus theorem \cite{motzkin1965maxima}, the global maximum of (\ref{optproblem1}) corresponds to the maximum clique on $G$. Thus, we can run both discrete replicator dynamic and path following replicator dynamic on the same graph, to see which dynamic has a higher chance to find the maximum clique. Of course, both dynamics may fail to find the maximum clique, since the maximum clique problem is a notoriously hard problem \cite{bomze1999maximum}.

To get statistically meaningful results, we need a large number of graphs with known maximum cliques. Thus, we choose to randomly generate graphs. In our experiment, graph $G$ is generated in the following way. First, generate a clique $G_1$ of $m_1$ vertices. Second, randomly generate a graph $G_2$. $G_2$ has $m_2$ vertices and $\alpha\frac{m_2(m_2-1)}{2}$ edges, where $\alpha$ controls the density of this graph. Besides, $G_2$ needs to follow a certain kind of degree distribution. Four kinds of degree distributions, namely, uniform distribution (U), binomial distribution (B), geometric law distribution (G) and power law distribution (P), are tested. According to \cite{newman64random}, these are the most common degree distributions observed in real networks. Finally, we randomly add $\beta m_1 m_2$ edges between $G_1$ and $G_2$, to form the final graph $G$. Note that when both $\alpha$ and $\beta$ are not large, the probability that $G_1$ is the maximum clique of $G$ is extremely high.

In our experiment, $m_1=100$ and $m_2=900$, thus, $G$ has $1000$ vertices in total. We set $\alpha=0.11$ and $\beta=0.005$, to make the average degrees of $G_1$ and $G_2$ approximately the same. For each type of degree distribution, we randomly generate $100$ graphs. For path following replicator dynamic, we use three samplings of path parameter, that is, $\Phi_1=\{\frac{1}{900},\frac{1}{800},\ldots,\frac{1}{100},1\}$, $\Phi_2=\{\frac{1}{950},\frac{1}{900},\ldots,\frac{1}{50},1\}$ and $\Phi_3=\{\frac{1}{990},\frac{1}{980},\ldots,\frac{1}{10},1\}$. Recall that discrete replicator dynamic is a special case of path following replicator dynamic, with $\Phi=\{1\}$.

The experimental results are shown in Table $1$. Both the percentage of successfully detected the maximum cliques $G_1$ and the average running time are reported. Clearly, the path following replicator dynamic is much more robust than discrete replicator dynamic. Moreover, the denser the samples are, the more robust the evolution process is. This is consistent with our intuition. For the time complexity, generally speaking, more samples lead to increased time; however, the run time seems to be not linear, but sub-linear, in the number of samples. For the random graphs with geometric law degree distribution, from $\Phi_1$ to $\Phi_2$, the average run time surprisingly drops, although $\Phi_2$ contains much more samples than $\Phi_1$. One possible explanation is that as the number of samples increases, the number of  fixed point iterations under each value of $\varepsilon$
reduces.

\begin{table}[t]
\caption{The results of DRD and PFRD on the maximum clique problem, where $\%$ denotes the percentage of correctly detected maximum cliques.}
\centering
\scalebox{0.9}
{
\begin{tabular}{l|c|c|c|c|c|c|c|c}
\hline
\multirow{2}{*}{ }&
\multicolumn{2}{|c}{DRD} & \multicolumn{2}{|c}{PFRD($\Phi_1$)} & \multicolumn{2}{|c}{PFRD($\Phi_2$)} & \multicolumn{2}{|c}{PFRD($\Phi_3$)} \\
\hline
& $\%$ & Time & $\%$ & Time & $\%$ & Time & $\%$ & Time \\
\hline
U & $0$ & $0.046$ & $79$ & $0.087$ & $100$ & $0.151$ & $100$ & $0.420$\\
B & $100$ & $0.067$ & $100$ & $0.137$ & $100$ & $0.207$ & $100$ & $0.391$\\
G & $0$ & $0.074$ & $0$ & $0.171$ & $95$ & $0.130$ & $100$ & $0.337$\\
P & $0$ & $0.048$ & $0$ & $0.124$ & $91$ & $0.134$ & $100$ & $0.363$\\
\hline
\end{tabular}
}
\end{table}

\subsection{Densest $k$-Subgraph Problem}
\begin{figure*}[t]
\centering
\includegraphics[width=1\linewidth]{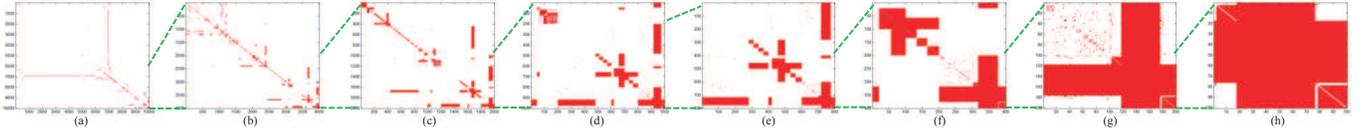}
\caption{Graphical illustration of the densest $k$-subgraphs of the web graph \emph{cnr-2000}, with $k$ being $10000$, $4000$, $2000$, $1000$, $800$, $400$, $200$ and $100$, respectively. Red color represents $1$, and white color represents $0$.
}
\end{figure*}

\begin{figure*}[t]
\centering
\includegraphics[width=1\linewidth]{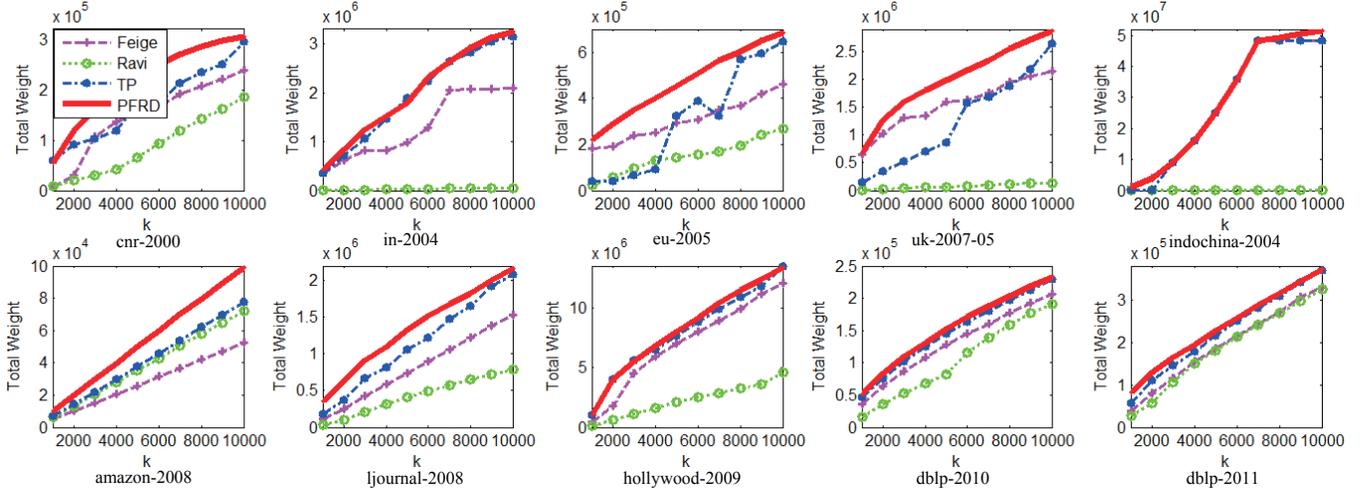}
\caption{ The results of D$k$S on $10$ large webgraphs. Feige's method is shown in magenta dashed curve, Ravi's method is shown in green dotted curve, the truncated power method is shown in blue dashdot curve, and our method is shown in red solid curve. This figure is best viewed in color.
}
\end{figure*}

As maximum clique problem, the densest $k$-subgraph problem (D$k$S) is also a fundamental and notoriously hard problem \cite{feige2001dense}. For a graph $G$, the task is to find a subgraph $G'$ with $k$ vertices, and the total weight of edges in $G'$ is the maximum among all subgraphs of $G$ with $k$ vertices. Algorithms for finding D$k$S are useful tools for many tasks. For example, they have been successfully used to select features for ranking \cite{geng2007feature}, to identify cores of communities \cite{kumar1999trawling}, and to combat link spam \cite{gibson2005discovering}.

Since D$k$S is an NP hard problem, many heuristic-based algorithms have been proposed. For general $k$,
the algorithm developed by Feige et al. \cite{feige2001dense} achieves the best approximation ratio of O($n^{\epsilon}$) with
$\epsilon<1/3$. Ravi et al. \cite{ravi1994heuristic} proposed $4$-approximation algorithms for weighted D$k$S problem on complete graphs for which the weights satisfy the triangle inequality. Recently, Yuan and Zhang have proposed truncated power method \cite{yuan2011truncated}, and achieved the state-of-the-art results on large web graphs.  In general, however, Khot \cite{khot2006ruling} showed that D$k$S has no polynomial time approximation scheme (PTAS), assuming that there are no sub-exponential time algorithms for problems in NP.

Suppose the solution of (\ref{optproblem2}) with $\varepsilon=\frac{1}{k}$ is $\mathbf{\tilde{x}}$, then $\mathbf{\tilde{x}}$ has at least $k$ positive components. The $k$ largest components in $\mathbf{\tilde{x}}$ form a set, denoted by $V_k$. Recall that $\mathbf{\tilde{x}}$ represents a cluster and $\tilde{x}_i$ is the probability of this cluster containing the vertex $v_i$, then the subgraph induced by $V_k$ is a natural candidate for the densest $k$-subgraph. An obvious advantage of our approach is: we can get the candidates of the densest $k$-subgraph for many $k$s in a single evolution process.

Fig. $3$ demonstrates the detected densest $k$-subgraphs on \emph{cnr-2000}. In one evolution process, dense subgraphs of various sizes are detected. As $k$ decreases, the obtained dense subgraph becomes denser and denser, finally it nearly becomes a whole graph when $k=100$. Obviously, the evolution process reveals important information of a graph and it is very useful in graph analysis.

We compare our approach with three methods, namely, Feige's method \cite{feige2001dense}, Ravi's method \cite{feige2001dense}, and truncated power method (TP)\cite{yuan2011truncated}. The source codes of these three methods are downloaded from web\footnote{https://sites.google.com/site/xtyuan1980}. For our approach, it is speeded up by setting $\delta_3=1e-12$, that is, setting the components of $\mathbf{x}(t)$ whose values are smaller than $1e-12$ to be zeros.

We use $10$ web graphs, namely, \emph{cnr-2000}, \emph{in-2004}, \emph{eu-2005}, \emph{uk-2007-05}, \emph{indochina-2004}, \emph{amazon-2008}, \emph{ljournal-2008},
\emph{hollywood-2009}, \emph{dblp-2010} and \emph{dblp-2011}. All these web graphs are from the WebGraph framework provided by the Laboratory for Web Algorithms\footnote{Datasets are available at http://lae.dsi.unimi.it/datasets.php}. For directed graphs, we treat each directed arc as an undirected edge. Table $2$ lists the statistics of the web graphs used in the experiment.

\begin{table}[t]
\caption{The statistics of the web graph datasets.}
\centering
\scalebox{1}
{
\begin{tabular}{l|c|c|c}
\hline
Graph & Vertices($|V|$) & Arc($|E|$) & Average Degree \\
\hline
\hline
\emph{cnr-2000} & $325,557$ & $3,216,152$ & $9.88$ \\
\emph{in-2004} & $1,382,908$ & $16,917,053$ & $12.23$ \\
\emph{eu-2005} & $862,664$ & $19,235,140$ & $22.30$ \\
\emph{uk-2007-05} & $1,000,000$ & $41,247,159$ & $41.25$ \\
\emph{indochina-2004} & $7,414,866$ & $194,109,311$ & $26.18$ \\
\emph{amazon-2008} & $735,323$ & $5,158,388$ & $7.02$ \\
\emph{ljournal-2008} & $5,363,260$ & $79,023,142$ & $14.73$ \\
\emph{hollywood-2009} & $1,139,905$ & $113,891,327$ & $99.91$ \\
\emph{dblp-2010} & $326,186$ & $1,615,400$ & $4.95$ \\
\emph{dblp-2011} & $986,324$ & $6,707,236$ & $6.80$ \\
\hline
\end{tabular}
}
\end{table}

\begin{table}[t]
\caption{Time used in D$k$S experiment. The
time is measured in seconds.}
\centering
\scalebox{1}
{
\begin{tabular}{l|c|c|c|c}
\hline
 Graph & Feige & Ravi & TP & PFRD \\
\hline
\hline
\emph{cnr-2000} & $1.87$ & $688.31$ & $8.77$ & $10.75$\\
\emph{in-2004} & $4.93$ & $1102.55$ & $23.75$ & $17.03$ \\
\emph{eu-2005} & $4.61$ & $1268.73$ & $39.57$ & $36.49$ \\
\emph{uk-2007-05} & $16.78$ & $2719.34$ & $50.70$ & $47.06$ \\
\emph{indochina-2004} & $31.06$ & $3531.42$ & $1321.69$ & $1257.47$ \\
\emph{amazon-2008} & $2.22$ & $999.93$ & $38.41$ & $12.12$ \\
\emph{ljournal-2008} & $30.67$ & $1588.95$ & $258.92$ &$325.11$ \\
\emph{hollywood-2009} & $15.41$ & $1531.01$ & $187.51$ &$251.36$ \\
\emph{dblp-2010} & $1.68$ & $3516.15$ & $9.08$ &$7.38$\\
\emph{dblp-2011} & $5.77$ & $1121.72$ & $33.58$ &$37.43$\\
\hline
\end{tabular}
}
\end{table}

For each graph, we compute its densest $k$-subgraphs for $10$ values of $k$, that is, $\{1000,2000,\ldots,10000\}$. Both total weights of obtained subgraphs and running time are reported. To save space, for each method, we only report its total time of obtaining $10$ subgraphs on each web graph. Our method compute all $10$ densest $k$-subgraphs of one graph in one evolution process; while other methods compute them separately.

Fig. $4$ shows the total weight of dense subgraphs versus the cardinality $k$. From
the performance curves, we can observe that our approach consistently outperform other three methods on all graphs. Truncated power method performs well on $5$ web graphs, but badly on \emph{eu-2005} and \emph{uk-2007-05}. Ravi's method performs worst on nearly all web graphs, except for \emph{amazon-2008}. Besides, our method performs extremely well for small $k$, this is because it obtains much more compact clusters than other methods.

The running time is reported in Table $3$. Three methods, Feige's method, truncated power method and our method, are efficient; while Ravi's method is time consuming. Feige's method is the fastest, since it only needs a few degree
sorting operations. Both truncated power method and our method need iterative matrix-vector multiplications, and they nearly have the same time complexity. Matrix-vector multiplication means the time complexity is linear in the number of edges, which is still very efficient. For example, on the web graph \emph{ljournal-2008}, which has $5,363,260$ vertices and $79,023,142$ edges, the whole evolution process of our approach only costs $325.11$ seconds. In fact, it can run much faster on a computer with larger memory. This is because the main limitation is the memory for large graphs. For example, due to insufficient memory, both truncated power method and our approach slow down on the web graph \emph{indochina-2004}.

\subsection{Structure Fitting}
\begin{figure*}[t]
\centering
\includegraphics[width=1\linewidth]{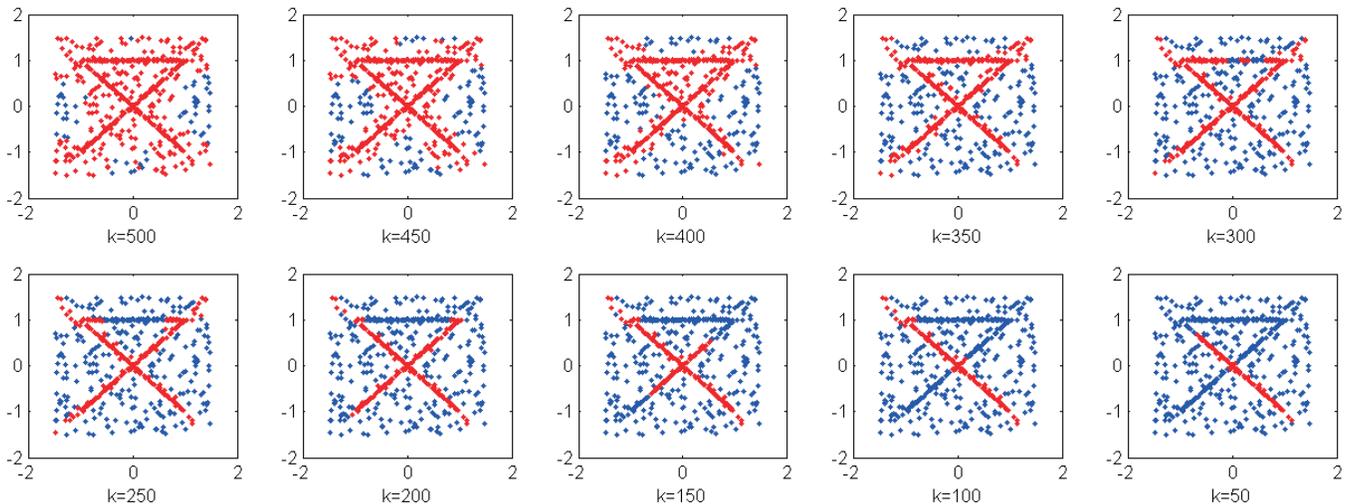}
\caption{Graphical illustration of path following replicator dynamic on a planar point cloud ($\varepsilon=\frac{1}{k}$). The point cloud has three lines, with each having $100$ inliers, and there are also $300$ outliers. Red points are inside the detected clusters, and blue points are outside the detected clusters.
}
\end{figure*}

Structure fitting, that is, fitting a geometric model to data, is a fundamental task in computer vision \cite{fischler1981random}. A typical example is to fit lines for a given point set. In structure fitting, two key questions are: $1$) how many structures in data? and $2)$ what are the parameters of these structures?
In practice, structure fitting is a non-trivial task, since real-world data may contain single or multiple structures, and may also be contaminated by severe noises and large amount of outliers.

At the year of $1981$, Fischler and Bolles proposed the seminar work for structure fitting, RANSAC \cite{fischler1981random}.
RANSAC utilizes a ``hypothesize-and-verify" framework to detect potential structures. This framework is very robust to outliers, which is the main challenge in structure fitting problems. After RANSAC, many structure fitting methods were proposed \cite{vedaldi2005,chum2008optimal,wang2012simultaneously}. Although these methods improve the performance in some aspects, basically, they all adopt the ``hypothesize-and-verify" framework of RANSAC.

The ``hypothesize-and-verify" framework generates many candidate structures, but does not tell us which of them are real structures.
In fact, among these candidate structures, most are fake structures, and some candidate structures correspond to the same real structures. Thus, we need to eliminate duplications and select real structures according to a certain criterion, or alternatively, estimate the number of real structures. In this direction, three representative works are J-linkage \cite{toldo2008robust}, KF \cite{chin2009robust} and RCG\cite{liu2012efficient}. In J-linkage\cite{toldo2008robust}, a ``conceptual representation", essentially a reduction of the parameter space to a one-dimensional discrete space of hypothesis index, is proposed. Robust fitting then proceeds by agglomerative clustering of the conceptual representations of the data points. In KF\cite{chin2009robust}, Chin proposed to sort the residuals and construct an affinity measure based on the sorted residuals. Such affinity measure can be used as kernel to estimate the number of clusters \cite{chin2009robust}. In RCG, we virtually construct a hypergraph, called random consensus graph (RCG), based on random sampling, then we build a binary graph from this virtual graph and obtain real structures by detecting dense subgraphs on the binary graph. The binary graph has nearly the same cluster structure as random consensus graph, but needs much less memory and can be constructed more efficiently.


We first do experiments on random planar point sets. The point set is generated as follows: first generate $3$ lines, with each line containing $n_i$ points, then add Gaussian noise $N(0,\sigma)$ to these points, finally add $n_o$ uniformly distributed outliers to the point set. The whole point set is within the region $[-1.5,1.5]^2$.

For line fitting, the basic relation is of order three, that is, for any three points, we can determine whether they are on a line or not. Thus, from the point set, we can build a hypergraph, where each hyperedge is of order three. However, it is time consuming to construct the hypergraph, and the hypergraph has so many hypereges as to fill up all memory quickly. Instead, we use the method proposed in \cite{liu2012efficient} to directly construct a binary graph, and run path following replicator dynamic on this binary graph.

Fig. $5$ illustrates the path following replicator dynamic on an exemplar point cloud. In this experiment, $n_i=100$, $n_o=300$ $\sigma =0.01$, $\delta_1=\frac{1}{600}$ and $1000$ hypotheses are generated. For each $\mathbf{\tilde{x}}$, the points in $C_{\mathbf{\tilde{x}}}$ are shown in red, others are shown in blue. As the figure shows, in the evolution process, outliers are dropped first, then points on one line, points on another line, and finally only points on the third line remain. Clearly, this process is very helpful for us to precisely detect all lines, since inliers of different lines are separated, as well as outliers.

To quantitatively measure the performance of this process, we randomly generate $100$ point sets, and calculate the average proportion of inliers in the cluster $C_{\mathbf{\tilde{x}}}$ for different values of $\varepsilon$. For each line, all the points whose distances to it are smaller than $\sigma=0.01$ are considered as inliers. The union of inliers of all three lines form the ground truth set $GT$. For each $\mathbf{\tilde{x}}$, we get a set $C_{\mathbf{\tilde{x}}}=\{i|x_i>\frac{1}{600}\}$. Then $\rho=\frac{|C_{\mathbf{\tilde{x}}}\cap GT|}{|C_{\mathbf{\tilde{x}}}|}$ is defined as precision of the set $C_{\mathbf{\tilde{x}}}$.

Fig. $6$ plots the average precision over $100$ experiments as a function of $k$, where $k=\frac{1}{\varepsilon}$. Both mean precision and one \emph{std} below the mean are illustrated. When $k$ is small, as expected, nearly all points in $C_{\mathbf{\tilde{x}}}$ are inliers. As $k$ increases, the precision drops a little, but still very high. This is because some points close to the three lines in data are considered as outliers. Finally, when $k$ is very large, outliers are involved and the precision drops fast. The \emph{std} is always small, which means that in the $100$ experiments, path following replicator dynamic performs steadily.

As Fig. $5$ shows, the sequence $\{C_{\mathbf{\tilde{x}}(1)},C_{\mathbf{\tilde{x}}(2)},\ldots,C_{\mathbf{\tilde{x}}(m)}\}$ gradually reveals the real structures in data, with points in the same structure gathering together. Thus,
by backtracking the sequence $\{C_{\mathbf{\tilde{x}}(1)},C_{\mathbf{\tilde{x}}(2)},\ldots,C_{\mathbf{\tilde{x}}(m)}\}$, we can reveal real structures one by one. The algorithm is summarized in Alg. $3$.

\begin{algorithm}[tb]
   \caption{Structure Fitting Based on Path Following Replicator Dynamic}
   \label{alg:gs}
\begin{algorithmic}[1]
   \STATE {\bfseries Input:} The cluster sequence $\{C_{\mathbf{\tilde{x}}(1)},C_{\mathbf{\tilde{x}}(2)},\ldots,C_{\mathbf{\tilde{x}}(m)}\}$, and a deviation threshold $\delta_4$.
   \STATE Fit a structure $s$ to $C_{\mathbf{\tilde{x}}(m)}$ and set $\digamma =\{s\}$, set $\daleth=C_{\mathbf{\tilde{x}}(m)}$;
   \FOR{$i=m-1,\ldots,1$}
   \STATE Set $\gimel = C_{\mathbf{\tilde{x}}(i)}/\daleth$.
   \FOR{Each vertex $v\in \gimel$}
    \STATE If the deviation of $v$ to any structure in $\digamma$ is smaller than a threshold $\delta_4$, then $\daleth=\daleth\cup{v}$ and $\gimel=\gimel/{v}$.
    \ENDFOR
   \IF{$|\gimel|\geq2*\lceil\frac{1}{\varepsilon_m}\rceil$}
   \STATE Fit a structure $s$ to the vertices in $\gimel$. All the vertices in $\gimel$ whose deviations to structure $s$ are smaller than $\delta_4$ form a set $\aleph$.
   \IF{$|\aleph|>\lceil\frac{1}{\varepsilon_m}\rceil$}
   \STATE Set $\digamma=\digamma\cup \{s\}$ and $\daleth=\daleth\cup \aleph$;
   \ELSE
   \STATE Terminate;
   \ENDIF
   \ENDIF
   \ENDFOR
   \STATE{\bfseries Output:} The set of fitted structures, $\digamma$.
\end{algorithmic}
\end{algorithm}

In Alg. $3$, $\digamma$ stores all the fitted structures, and $\daleth$ contains vertices which are inliers of these structures. For a structure $s$, its inliers are the vertices whose deviations to it are smaller than a threshold $\delta_4$.
Since we detect a new structure from the vertices in $\gimel$, we require that $\gimel$ contains a sufficient number of vertices. Here the threshold $2*\lceil\frac{1}{\varepsilon_m}\rceil$ is twice the least number of vertices in the cluster $C_{\mathbf{\tilde{x}}(m)}$. If the fit of structure $s$ to the vertices in $\gimel$ does not have enough inliers, then the algorithm terminates. The same as many structure fitting methods, Alg. $3$ is controlled by two parameters, the deviation threshold $\delta_4$ and the minimal number of inliers in a structure, $\lceil\frac{1}{\varepsilon_m}\rceil$. Note that duplications are automatically eliminated in the backtracking process of Alg. $3$.

We compare Alg. $3$ with three methods, namely, J-Linkage, KF and RCG. J-linkage partitions data into many clusters, and large clusters are regarded as real structures. For KF, it has no parameter, since it can automatically estimate the number of clusters by some heuristic rules; however, the estimated number may be incorrect sometimes, this is because estimating the number of clusters in data is a notoriously hard problem\cite{tibshirani2001estimating}. For RCG, it considers clusters with large objective values as real structures.

We test all four methods on randomly generated point sets, under different levels of noises. We keep $n_i=100$ and $n_o=300$, and increase the noise parameter $\sigma$ from $0.01$ to $0.08$, with step $0.01$. For each point set, we randomly select $1000$ minimal size samples and generate $1000$ hypotheses. For each value of $\sigma$, we repeat the experiments $100$ times. The performance is measured by \emph{average fitting error}, that is, the average distance of inliers of three lines to their corresponding fitted lines. Here inliers mean points whose distances to any of three real lines are smaller than $\sigma$.
For each method, we tune its parameters to obtain the best performance\footnote{This may be a little unfair for KF method, since it has no parameter to tune.}.  The results are reported in Table $4$, both average fitting error and average running time (in parentheses) are reported. For KF, the estimated number of structures may be incorrect, thus, we also report the times when it correctly estimates the number of structures (in square brackets), and its average fitting error is averaged only over these experiments. Both RCG and our method have much better performance, since the structures are estimated from points in dense clusters, which are mostly points on the real lines. The performance of J-linkage degrades quickly as the level of noises increases, probably because the obtained cluster is not so compact. For the time complexity, KF is most time consuming, since it does singular value decomposition on kernel matrix. J-linkage is also slow, due to its agglomerative clustering step. RCG is the fastest, since its computation is restricted to small subgraphs. Our method is only a bit slower than RCG. This is because the evolution process operates on the whole graph at the first few iterations.

\begin{figure}[t]
\centering
\includegraphics[width=1\linewidth]{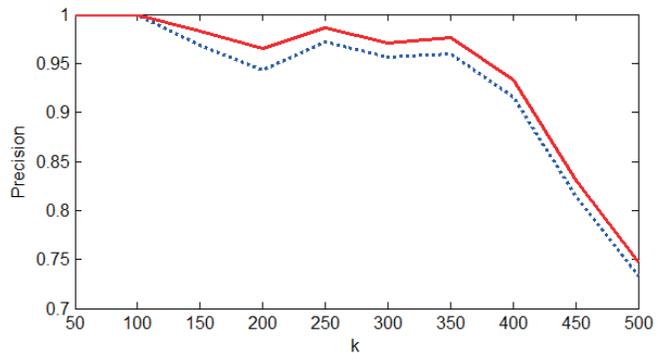}
\caption{The proportion of inliers in the clusters of path following replicator dynamic. Red solid curve illustrates the average precision, and blue dotted curve illustrates the curve of one \emph{std} below the mean.
}
\end{figure}

\begin{figure}[t]
\centering
\includegraphics[width=1\linewidth]{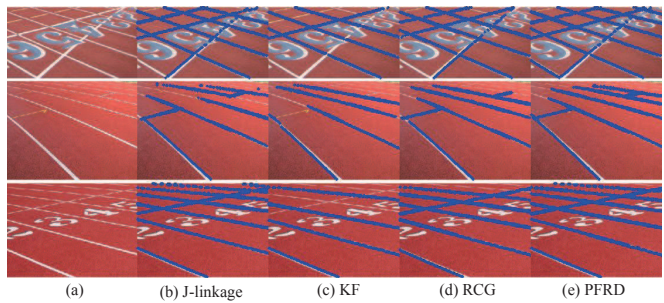}
\caption{Results of line fitting on three real images.
}
\end{figure}

In Fig. $7$, we illustrate the results of line fitting on three real images. In each image, there are multiple lines, and the sizes of these lines vary drastically. It is a difficult task to detect short lines without false alarm. This is because some fake structures are ranked higher than real short lines. Thus, for each method, we only illustrate the real structures in its top-$k$ ranked detections, with $k$ being $12$, $10$, $10$ for the first, second and third image\footnote{For KF method, we report the real structures in all of its detections.}, respectively. Note that for some tracks, both of its two edges are detected. For clarity, we only illustrate one of them. Generally speaking, all four methods successfully detect long lines, and the differences lie in the detection of short lines. Due to the help of evolution process, our method correctly detect most of short lines. KF fails to detect all short lines, probably because it regards them as outliers, thus estimates the number of clusters incorrectly. Thus, this experiment also shows the difficulty of estimating the number of clusters in real data, especially for data with multiple scales.

\begin{table}[t]
\caption{Experimental results on line fitting. In each cell, the top value is the average fitting error, the bottom value is the average run time, measured in seconds.}
\centering
\scalebox{0.8}
{
\begin{tabular}{l|c|c|c|c}
\hline
 $\sigma$ & J-Linkage & KF & RCG & Our method\\
\hline
\hline
\multirow{2}{*}{$0.01$} & $9.449e-4$ & $9.462e-4[83]$ & $9.634e-4$ & $\mathbf{8.585e-4}$ \\
& ($\num{0.5132}$) & ($\num{0.7386}$) & ($\num{0.1202}$) & ($\num{0.1315}$) \\
\hline
\multirow{2}{*}{$0.02$} & $0.0037$ & $0.0031 [67]$ & $0.0025$ & $\mathbf{0.0023}$ \\
& ($0.5607$) & ($0.7785$) & ($0.1198$) & ($0.1386$) \\
\hline
\multirow{2}{*}{$0.03$} & $0.0065$ & $0.0053 [75]$ & $0.0040$ & $\mathbf{0.0038}$ \\
& ($0.5186$) & ($0.7682$) & ($0.1172$) & ($0.1357$) \\
\hline
\multirow{2}{*}{$0.04$} & $0.0146$ & $0.0097[72]$ & $0.0068$ & $\mathbf{0.0067}$ \\
& ($0.5860$) & ($0.7841$) & ($0.1110$) & ($0.1204$) \\
\hline
\multirow{2}{*}{$0.05$} & $0.0178$ & $0.0138 [54]$ & $0.0076$ & $\mathbf{0.0069}$ \\
& ($0.5339$) & ($0.7771$) & ($0.1102$) & ($0.1143$) \\
\hline
\multirow{2}{*}{$0.06$} & $0.0217$ & $0.0174 [52]$ & $\mathbf{0.0118}$ & $0.0122$ \\
& ($0.5678$) & ($0.7634$) & ($0.1151$) & ($0.1247$) \\
\hline
\multirow{2}{*}{$0.07$} & $0.0245$ & $0.0196 [58]$ & $\mathbf{0.0124}$ & $0.0129$ \\
& ($0.5882$) & ($0.7584$) & ($0.1142$) & ($0.1264$) \\
\hline
\multirow{2}{*}{$0.08$} & $0.0372$ & $0.0251 [47]$ & $0.0155$ & $\mathbf{0.0151}$ \\
& ($0.5417$) & ($0.7930$) & ($0.1158$) & ($0.1253$) \\
\hline
\end{tabular}
}
\end{table}

\subsection{Discovery of High-Density Regions}
Give a dataset $P$ with $n$ points, $P=\{p_1,\ldots,p_n\}$, we can construct a graph $G$, with the weight $w_{ij}=\exp(-\frac{d^2(p_i,p_j)}{h^2})$, where $d(p_i,p_j)$ represents the distance between $p_i$ and $p_j$, and $h$ is the bandwidth parameter. As mentioned before, the path following replicator dynamic on such graph $G$ can be considered as a shrink process of high-density regions. In this shrink process, high-density regions at different scales, which represent important patterns in dataset $P$, will naturally emerge.

\begin{figure}[t]
\centering
\includegraphics[width=1\linewidth]{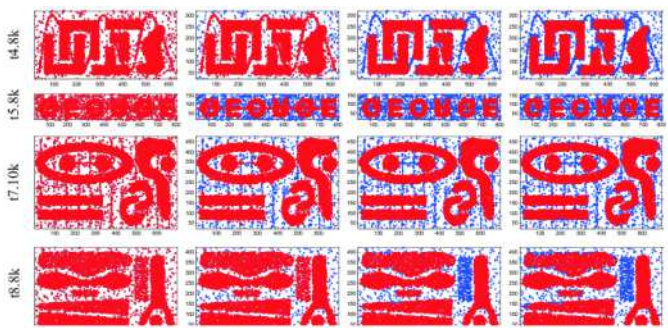}
\caption{The evolution process of path following replicator dynamic on Chameleon dataset \cite{karypis1999chameleon}. The points in clusters are shown in red; while the points out of clusters are shown in blue. Clearly, as $\varepsilon$ increases, outliers gradually disappear and high-density regions emerge.
}
\end{figure}

Fig. $8$ demonstrates the evolution process of path following replicator dynamic on Chameleon dataset \footnote{http://glaros.dtc.umn.edu/gkhome/cluto/cluto/download}, which consists four $2$d point sets. These four point sets contain high-density regions in complex forms, as well as outliers. The second point set contains $10000$ points, and other three point sets contain $8000$ points. In this experiment, we set $h=10$ and $\delta_1=0.0001$.
As the results illustrate, path following replicator dynamic can successfully eliminate outliers and reveal high-density regions, despite there are multiple disjoint high-density regions and the shape of these high-density regions are complex, which is a big challenge to many other methods, such as mean shift \cite{comaniciu2002mean} and one-class SVM \cite{scholkopf2001estimating}.

We also do experiment on a hand pose dataset, which is collected by a human-computer interaction software, using Micorsoft Kinect\footnote{http://www.microsoft.com/en-us/kinectforwindows/}. The user interacted with computer by hand poses, with some poses having specific meanings, and others being meaningless.
The dataset contains $12000$ instances of hand poses in total. Each instance is a $80\times 80$ depth image. There are three meaningful hand poses, namely, \emph{extend}, \emph{point} and \emph{fist}, and each has $2000$ instances. The other $6000$ instance are meaningless, thus are considered as outliers. The task is to discover meaningful hand poses, and at the same time, identify outliers. Each meaningful hand pose may be captured in different viewpoints and distances, thus all of its instances form a high-density region of complex shape. We compare with three methods, namely, mean shift (MS) \cite{comaniciu2002mean}, ensemble clustering (EC) \cite{liu2010affinity} and one-class SVM ($1$-SVM)\cite{CC01a}. For our method, we set $\Psi=\{\frac{1}{10000},\frac{1}{9000},\ldots,\frac{1}{6000}\}$. According to $\mathbf{\tilde{x}}(5)$ ($\varepsilon=\frac{1}{6000}$), we obtain a cluster of $6000$ points, then compute its \emph{precision}, which is the proportion of inliers in this cluster. We also correctly estimate the number of meaningful poses on this cluster by gap statistic \cite{tibshirani2001estimating}. For mean shift, it detects a large number of modes, and in a high-density region, there are usually multiple modes. To identify outliers, we choose $3$ most significant modes, and for each mode, we find its $2000$ nearest neighbors. In this way, we also get a cluster of $6000$ points and compute its precision. In ensemble clustering, there is a parameter to control the size of each detected dense cluster, and we set it to $2000$. Three significant clusters, with each containing $2000$ points, are detected. We compute the precision of the union of these three clusters. For one-class SVM \footnote{we use libsvm: http://www.csie.ntu.edu.tw/\~cjlin/libsvm}, it calculates a surface to separate inliers and outliers. We adjust its parameter to make the numbers of points one both sides of surface are approximately equal, then calculate the precision of points on the side classified as inliers. The precision of all four methods is reported in Table $5$.

\begin{table}[t]
\caption{Experimental results on the hand pose dataset.}
\centering
\scalebox{1}
{
\begin{tabular}{l|c|c|c|c}
\hline
 & MS & EC & $1$-SVM & PFRD\\
\hline
Precision & $31.22\%$ & $75.28\%$ & $72.90\%$ & $\mathbf{85.62\%}$\\
\hline

\hline
\end{tabular}
}
\end{table}

As the experimental results show, PFRD significantly outperforms the other methods, because it identifies outliers by finding globally densest regions. Mean shift detects the densest points in feature space, which only indicates the existence of high-density regions. Since we simply use the distances to these high-density points to identify outliers, and the shape of high-density regions is complex, mean shift performs badly on this dataset. For ensemble clustering, since the least number of vertices in a cluster is set to $2000$, it detects high-density regions in the large scale, thus performs well. Due to the usage of kernel trick, one-class SVM can detect high-density regions of complex shapes, and it also performs well one this dataset.

\section{Conclusion}
The proposed path following replicator dynamic is a generalization of discrete replicator dynamic.
The introduced dynamic path parameter controls the behavior of the evolution process.
As a result, the evolution process is less sensitive to degree distribution and mainly determined by the global
structure of a graph. Due to its global awareness, the proposed
dynamic can automatically gather vertices based on their cluster membership. This makes it extremely
powerful and useful as a general tool for discovering the cluster structure
of graphs. This fact is demonstrated on four different applications.
Due to its high efficiency, the proposed method can be easily integrated
into many complex systems as a computationally cheap but effective module.


\ifCLASSOPTIONcaptionsoff
  \newpage
\fi

\bibliography{Ref}
\bibliographystyle{IEEETran}

\end{document}